\documentclass{llncs}
\pdfoutput=1
\usepackage[T1]{fontenc}

\usepackage{makeidx}

\usepackage{makeidx}
\usepackage{graphicx}
\usepackage{amssymb}
\usepackage{amsmath}
\usepackage{amscd}

\newcommand{\commentout}[1]{}

\newcommand{\introskip}{\vspace{3ex}}

\newcommand{\E}{\mathbb{E}}                    

\newcommand{\abs}[1]{\mathop{\left\lvert #1 \right\rvert}} 
\newcommand{\args}[1]{\mathop{\left( #1 \right)}} 
\newcommand{\inner}[1]{\mathop{\left\langle #1 \right\rangle}}
\newcommand{\norm}[1]{\mathop{\left\lVert #1 \right\rVert}}
\newcommand{\cbrace}[1]{\mathop{\left\{ #1 \right\}}}
\newcommand{\bracket}[1]{\mathop{\left[ #1 \right]}}

\newcommand{\argsS}[2]{\mathop{\left( #1 \right)#2}} 

\newcommand{\normS}[2]{\mathop{\left\lVert #1 \right\rVert#2}}

\newcommand{\T}{\mathop{\mathsf{T}}}           

\renewcommand{\S}[1]{{\mathcal{#1}}}           
\def\vec#1{\mathchoice{\mbox{\boldmath$\displaystyle#1$}}
{\mbox{\boldmath$\textstyle#1$}}
{\mbox{\boldmath$\scriptstyle#1$}}
{\mbox{\boldmath$\scriptscriptstyle#1$}}}

\newcommand{\atab}[1]{\hspace*{#1em}}

\newcounter{algorithm_counter}
\newenvironment{algorithm}[1]{
\refstepcounter{algorithm_counter}
\setlength{\parindent}{0\parindent}
\vspace{2ex}
\begin{minipage}{\textwidth}
\rule{\textwidth}{5\arrayrulewidth}\\
\begin{footnotesize}
{\bf Algorithm \arabic{algorithm_counter}} (#1) \\
\rule[+1.5ex]{\textwidth}{\arrayrulewidth}

\vspace{-1.5ex}

}%
{
\\[-1.5ex]
\rule{\textwidth}{\arrayrulewidth}
\end{footnotesize}
\end{minipage}
\setlength{\parindent}{\parindent}
}

{
\\[-1.5ex]
\rule{\textwidth}{\arrayrulewidth}
\end{footnotesize}
\end{minipage}
\setlength{\parindent}{\parindent}
}

\begin{document}
\title{Elkan's k-Means for Graphs}
\titlerunning{Elkan's k-Means Algorithms for Graphs}

\author{Brijnesh J.~Jain$^1$  \and Klaus Obermayer$^1$} 
\authorrunning{Brijnesh J.~Jain and Klaus Obermayer}  
\institute{$^1$Berlin Institute of Technology, Germany\\
\email{\{jbj|oby\}@cs.tu-berlin.de}}

\maketitle 

\begin{abstract}  
This paper extends k-means algorithms from the Euclidean domain to the domain of graphs. To recompute the centroids, we apply subgradient methods for solving the optimization-based formulation of the sample mean of graphs. To accelerate the k-means algorithm for graphs without trading computational time against solution quality, we avoid unnecessary graph distance calculations by exploiting the triangle inequality of the underlying distance metric following Elkan's k-means algorithm proposed in \cite{Elkan03}. In experiments we show that the accelerated k-means algorithm are faster than the standard k-means algorithm for graphs provided there is a cluster structure in the data.
\end{abstract}

\section{Introduction}

The k-means algorithm is a popular clustering method because of its simplicity and speed. The algorithmic formulation of k-means as well as the solutions of its cluster objective presuppose the existence of a sample mean. Since the concept of sample mean is well-defined for vector spaces only, application of the k-means algorithm has been limited to patterns represented by feature vectors. But often, the objects we want to cluster have no natural representation as feature vectors and are more naturally represented by finite combinatorial structures such as, for example, point patterns, strings, trees, and graphs arising from diverse application areas like proteomics, chemoinformatics, and computer vision.  

For combinatorial structures, pairwise clustering algorithms are one of the most widely used methods to partition a given sample of patterns, because they can be applied to patterns from any distance space without any  additional mathematical structure. Related to k-means, the k-medoids algorithm is a well-known alternative that can also be applied to patterns from an arbitrary distance space. The k-medoids algorithm operates like k-means, but replaces the concept of mean by the set median of a cluster \cite{Schenker03}. With the emergence of the generalized median \cite{Bunke01,Ferrer07} and sample mean of graphs \cite{Jain04,Jain08}, variants of the k-means algorithm have been extended to the domain of graphs \cite{Jain04,Ferrer07,Jain08}. 

In an unmodified form, however, pairwise clustering, k-medoids as well as the extended k-means algorithm are slow in practice for large datasets of graphs. The main obstacle is that determining a graph distance is well known to be a graph matching problem of exponential complexity. But even if we resort to graph matching algorithms that approximate graph distances in polynomial time, application of  clustering algorithms for large datasets of graphs is still hindered by their prohibitive computational time. 

For pairwise clustering, the number of NP-hard graph distance calculations depends quadratically on the number of the input patterns. In the worst-case, when almost all patterns are in one cluster, k-medoids also has quadratic complexity in the number of distance calculations. If the $N$ graph patterns are uniformly distributed in $k$ clusters, k-medoids requires $O(tN^2\!/k)$ graph distance calculations, where $t$ is the number of iterations required. For k-means, we require $kN$ graph distance calculations at each iteration in order to assign $N$ pattern graphs to their closest centroids. Recomputing the centroids requires additional graph distance calculations. In the best case, when using the incremental arithmetic mean method \cite{Jain09b} for approximating a sample mean, $N$ graph distance calculations  at each iteration are necessary to recompute the centroids. This gives a total of  $t(k+1)N$ graph distance calculations, where $t$ is the number of iterations required. In view of the exponential complexity of the graph matching problem, reducing the number of distance calculations in order to make k-means for graphs applicable is imperative.

In this contribution, we propose an accelerated version of k-means for graphs by extending Elkan's method \cite{Elkan03} from vector to graphs.  For this we assume that the underlying graph distance is a metric. To avoid computationally expensive graph distance calculations, we exploit the triangle inequality by keeping track of upper and lower bounds between input graphs and centroids. 

The k-means algorithm for graphs generalizes the standard k-means algorithm for vectors. Regarding feature vectors as graphs consisting of a single attributed node, k-means for graphs coincides with k-means for vectors. The proposed accelerated version of k-means for graphs has the following properties: First, based on the $\S{T}$-space framework, accelerated k-means can be applied to finite combinatorial structures other than graphs like, for example, point patterns, sequences, trees, and hypergraphs. For sake of concreteness, we restrict our attention exclusively to the domain of graphs. Second, any initialization method that can be used for k-means for graphs can also be used for the Elkan's k-means for graphs. Third, k-means for graphs and its accelerated version perform comparable with respect to solution quality. Different solutions are due to the approximation errors of the graph matching algorithm and the non-uniqueness of the sample mean of graphs but are not caused by the mechanisms to accelerate the clustering algorithm. 

The paper is organizes as follows. Section 2 briefly describes the standard k-Means algorithm for vectors. Section 3 extends the standard k-means from vectors to graphs. Section 4 introduces Elkan's k-means algorithm for graphs. Experimental results are presented and discussed in Section 5. Finally, Section 6 concludes with a summary of the main results and future work. 

\section{The k-Means Algorithm for Euclidean Spaces}

This section describes k-means for vectors \cite{Theodoridis09} in order to point out commonalities and differences with k-means for graphs. 

\introskip

Suppose that we are given a training sample $\S{S} = \cbrace{\vec{x}_{1}, \ldots, \vec{x}_{N}}$ of $N$ vectors drawn from the Euclidean space $\S{X}$. A partition $\S{P} = \cbrace{\S{C}_1, \ldots, \S{C}_k}$ of $\S{S}$ into $k$ disjoint subsets (clusters) $\S{C}_i \subseteq \S{S}$ is determined by a $(N \times k)$-membership matrix $\vec{M} = (m_{ij})$ satisfying the constraints
\begin{align*}
\sum_{j = 1}^{k} m_{ij} = 1 &\quad \mbox{for all } i \in \cbrace{1, \ldots, N}\\
m_{ij} \in \cbrace{0, 1}   &\quad \mbox{for all } i \in \cbrace{1, \ldots, N}, j \in \cbrace{1, \ldots, k}.
\end{align*}
The standard k-means clustering algorithm aims at finding $k$ centroids $\S{Y} = \cbrace{\vec{y}_{1}, \ldots, \vec{y}_{k}}\subseteq \S{X}$ and a partition $\vec{M} = (m_{ij})$ of the set $\S{S}$ such that the cluster objective 
\[
J\args{\vec{M},\S{Y} \,|\, \S{S}} = \frac{1}{m}\sum_{j=1}^{k}\sum_{i=1}^{N} m_{ij} \normS{\vec{x}_{i} - \vec{y}_{j}}{^2},
\]
is minimized. 

Suppose that we fix an arbitrary membership matrix $\vec{M}$. Then the cluster objective $J(.\,|\, \vec{M}, \vec{X})$ given $\vec{M}$ and $\S{X}$ is differentiable as a function of the centroids $\S{Y}$. The $k$ centroids that minimize $J(.\,|\, \vec{M}, \vec{X})$ are the sample means
\[
\vec{y}_j = \frac{1}{\abs{\S{C}_j}}\sum_{i=1}^{N} m_{ij}\vec{x}_i, 
\]
of the clusters $\S{C}_j$ consisting of data points $\vec{x} \in \S{S}$ assigned to centroid $\vec{y}_j$. Since the $k$ sample mean centroids together with the given membership matrix $\vec{M}$ yields a local minimum of the cluster objective $J$ only, the challenging task of minimizing $J$ consists in finding an optimal membership matrix. Since this problem is NP-complete \cite{Garey82}, several heuristic algorithms have been devised. A standard clustering heuristic that minimizes $J$ is the k-means algorithm as outlined in Algorithm \ref{alg:k-means}. The notation $\S{C}(\vec{y})$ used in Algorithm \ref{alg:k-means} denotes the cluster associated with centroid $\vec{y} \in \S{Y}$.

\begin{figure}[t]
\centering
\begin{algorithm}{K-Means Algorithm for Euclidean Spaces}\label{alg:k-means}
01 \atab{1} choose initial centroids $\vec{y}_{1}, \ldots, \vec{y}_{k}\in \S{X}$\\
02 \atab{1} \textbf{repeat} \\
03 \atab{3} assign each $\vec{x} \in \S{S}$ to its closest centroid 
$\vec{y}_{\vec{x}} = \arg\min_{\vec{y}\in\S{Y}} \normS{\vec{x} - \vec{y}}{^2}$\\
04 \atab{3} recompute each centroid $\vec{y}\in \S{Y}$ as the mean of all vectors from $\S{C}(\vec{y})$\\
05 \atab{1} \textbf{until} some termination criterion is satisfied
\end{algorithm}
\end{figure}

\section{The k-Means Algorithm for Graphs}

To extend k-means from the domain of feature vectors to the domain of graphs, two modifications are necessary \cite{Jain04,Jain08}: First, we replace the Euclidean metric by a graph metric. Second, we replace the sample mean of vectors by a related concept for graphs. 

\subsection{Metric Graph Spaces}

In principle, we can substitute any graph metric into the standard k-means algorithm in order to obtain its structural counterpart. Here, we focus on geometric graph distances that are related to the Euclidean metric, because the Euclidean metric is the underlying metric of the vectorial mean. The vectorial mean in turn provides a link to deep results in probability theory and is the foundation for a rich repository of analytical tools in pattern recognition. To access at least parts of these results, it seems to be reasonable to relate graph metrics to the Euclidean metric. This restriction is acceptable from an application point of view, because geometric distance functions on graphs and their related similarity functions are a common choice of proximity measure  \cite{Almohamad93,Cour06,Gold96,Holm93,Umeyama1988,Wyk02}. 

Though it is straightforward to define a graph metric, which is related to the Euclidean metric of vectors, we first make a detour via the concept of $\S{T}$-space in order to approach the sample mean of graphs in a principled way. 

\introskip

Let  $\E$ be a $d$-dimensional Euclidean vector space.  An
(\emph{attributed}) graph is a triple $X = (V,E,\alpha)$ consisting of a
finite nonempty set $V$ of \emph{vertices}, a set $E \subseteq V \times V$
of \emph{edges}, and an \emph{attribute function} $\alpha:V\times V
\rightarrow \E$, such that $\alpha(i,j) \neq \vec{0}$ for each edge and 
$\alpha(i,j) = \vec{0}$ for each non-edge. Attributes $\alpha(i,i)$ of
vertices $i$ may take any value from $\E$. 

For simplifying the mathematical treatment, we assume that all graphs are of order $n$, where $n$ is chosen to be sufficiently large. Graphs of order less than $n$, say $m < n$, can be extended to order $n$ by including isolated vertices with attribute zero. For practical issues, it is important to note that limiting the maximum order to some arbitrarily large number $n$ and extending smaller graphs to graphs of order $n$ are purely technical assumptions to simplify mathematics. For machine learning problems, these limitations should have no practical impact, because neither the bound $n$ needs to be specified explicitly nor an extension of all graphs to an identical order needs to be performed. When applying the theory, all we actually require is that the graphs are finite. 

A graph $X$ is completely specified by its \emph{matrix representation}
$\vec{X} = (\vec{x}_{ij})$ with elements $\vec{x}_{ij} = \alpha(i,j)$ for
all $1 \leq i,j\leq n$. By  concatenating the columns of 
$\vec{X}$, we obtain a \emph{vector representation} $\vec{x}$ of $X$. 

Let $\S{X} = \E^{n \times n}$ be the Euclidean space of all ($n \times
n$)-matrices and let $\S{T}$ denote a subset of the set $\S{P}^n$ of all 
$(n\times n)$-permutation matrices. Two matrices
$\vec{X}\in\S{X}$ and $\vec{X'}\in\S{X}$ are said to be equivalent, if there
is a permutation matrix $P \in \S{T}$ such that $\vec{P}^{\T}\vec{X}\vec{P}
= \vec{X'}$. The quotient set 
\[
\S{X_T} = \S{X}/\S{T} = \cbrace{[\vec{X}] \,:\, \vec{X} \in \S{X}}
\] 
is the $\S{T}$-\emph{space} over the \emph{representation space} $\S{X}$. A $\S{T}$-space is a relaxation of the set $\S{G_T} = \S{G}/\S{T}$ of all \emph{abstract graphs} $[\vec{X}]$, where $\vec{X}$ is a matrix representation of graph $X$. 

In the remainder of this contribution, we identify $\S{X}$ with $\E^N$ ($N = n^2$) and consider vector- rather than matrix representations of abstract graphs. By abuse of notation, we sometimes identify $X$ with $[\vec{x}]$ and write $\vec{x} \in X$ instead of $\vec{x} \in \bracket{\vec{x}}$.  

Finally, we equip a $\S{T}$-space with a metric related to the Euclidean metric. Suppose that $d(\vec{x}, \vec{y}) = \norm{\vec{x} - \vec{y}}$ is an Euclidean metric on $\S{X}$ induced by some inner product. Then the distance function 
\[
D(X, Y) = \min \cbrace{d(\vec{x}, \vec{y}) \,:\, \vec{x} \in X, \vec{y} \in Y}
\] 
is a metric with the same geometric properties as $d$. A pair $(\vec{x}, \vec{y}) \in X \times Y$ of vector representations is called \emph{optimal alignment} if $D(X, Y) = d(\vec{x}, \vec{y})$. 

\subsubsection*{Calculating a Graph Metric.}  Here, we assume that $\S{T}$ is equal to the set of all $\S{P}^n$ of all $(n\times n)$-permutation matrices.
Determining a graph distance $D(X, Y)$ and finding an optimal alignment of $X$ and $Y$ are equivalent problems that are more generally referred to as a graph matching problem. In contrast to calculating the Euclidean distance between vectors, computing $D(X, Y)$ is a NP-complete problem \cite{Gold96}. Devising graph matching algorithms for computing $D(X, Y)$ has become a mature field in structural pattern recognition that has produced various powerful and efficient solutions to the graph matching problem \cite{Conte04}. To extend k-means to the graph domain any of those algorithms can be used. 

\subsection{The Sample Mean of Graphs}

Given the metric space $(\S{X_T}, D)$, we introduce the sample mean of graphs and provide some results proved in \cite{Jain08}. 

\introskip

Suppose that $\S{S_T} = \left(X_{1}, \ldots, X_{N}\right)$ is a sample of $m$ abstract graphs from $\S{G_T} \subseteq \S{X_T}$. A \emph{sample mean} of $\S{S_T}$ is any solution of the optimization problem   
\[
\args{P}\quad \begin{array}{l@{\quad}l}
  \mbox{min}& F(X) =  {\displaystyle\frac{1}{2}}{\displaystyle\sum_{i=1}^{N}  D(X, X_{i})^{2}}\\[1ex]
  \mbox{s.t.}&  X \in \S{X_T}
\end{array}.
\]
The cost function $F$ is the \emph{sum of squared distances} (SSD) to the sample graphs. Here, the problem is to find a
solution from an uncountable infinite set $\S{X_T}$. A simpler problem is to restrict the set $\S{X_T}$ of feasible solutions to the finite sample $\S{S_T} \subseteq \S{X_T}$. A \emph{set mean graph} of $\S{S_T}$ is
defined by 
\[
Y = \arg\min \cbrace{F(X) \,:\, X \in \S{S_T}}.
\]
We summarize the most important results from \cite{Jain08} for deriving subgradient-based algorithms for solving problem ($P$). 
\begin{theorem}\label{theorem:sample-mean}
Let $\S{S_T} = \left(X_{1}, \ldots, X_{N}\right) \subseteq \S{G_T}$ be a sample of $m$ abstract graphs. 
\begin{enumerate}
\item Problem ($P$) has a solution. The solutions are abstract graphs from $\S{G_T}$.
\item The SSD function $F$ is locally Lipschitz.    
\item A vector representation $\vec{y}$ of a sample mean $Y \in \S{X_T}$ of $\S{S_T}$ is of the form 
\[
\vec{y} = \frac{1}{N}\sum_{i=1}^{N}\vec{x}_i,
\] 
where $d(\vec{x}_i, \vec{y}) = D(X_i, Y)$ for all $i \in \cbrace{1,\ldots, N}$. We call the vector representations $\args{\vec{x}_1, \ldots, \vec{x}_N}$ an \emph{optimal multiple alignment} of $\S{S_T}$. 
\item Let $\args{\vec{x}_1, \ldots, \vec{x}_N}$ be an optimal multiple alignment of  $\S{S_T}$. Then 
\[
\sum_{i=1}^{N}\sum_{j = i+1}^{N}  \inner{\vec{x}_i, \vec{x}_j} 
\geq \sum_{i=1}^{N}\sum_{j = i+1}^{N}  \inner{\vec{x}'_i, \vec{x}'_j}
\]
for all vector representations $\vec{x}'_1 \in X_1, \ldots, \vec{x}'_N  \in X_N$. 
\end{enumerate}
\end{theorem}
The first statement ensures that problem ($P$) can be solved and has feasible solutions. Since the SSD satisfies the locally Lipschitz condition according to the second statement, we can apply generalized gradient techniques from nonsmooth optimization for minimizing the SSD \cite{Maekelae92}. The third statement shows that a vector representation of a structural sample mean is the standard sample mean of certain vector representations of the sample graphs. In addition, we see that problem ($P$) is a discrete rather than a continuous optimization problem, where a solution can be chosen from the finite set $X_1 \times \cdots \times X_m = \cbrace{\args{\vec{x}_1, \ldots, \vec{x}_m} \,:\, \vec{x}_i \in X_i}$. The latter property combined with the
fourth statement can be exploited for constructing search algorithms or
meta-heuristics like genetic algorithms. The fourth statement asks for
maximizing the sum of pairwise similarities (SPS). The standard sample mean
of a vector representation maximizing the SPS is a vector representation
of a structural sample mean. Apart from this, the fourth property provides a
geometric characterization stating that an optimal multiple alignment has
minimal volume within the subspace spanned by the vector representations. In
the case that $D$ is derived from the maximum common subgraph problem, the
fourth property says that an optimal multiple alignment maximizes the sum of
common edges of the sample graphs. This in turn indicates that computation
of the sample mean has potential applications in frequent substructure
mining.

\subsubsection*{A Subgradient Method for Approximating a Sample Mean.}

So far, we have defined a concept of sample mean for graphs. For practical applications, we need an efficient procedure to minimize problem ($P$) in order to recompute the centroids of the k-means algorithm for graphs. For this, we assume that $\S{S_T} = \args{X_1, \ldots, X_N}$  is a sample $\S{S_T} = \args{X_1, \ldots, X_N}$ of $m$ graphs.

\paragraph*{ Generic Subgradient Method.}
Suppose that we want to minimize a locally Lipschitz function $f$ on
$\S{X}$. Then $f$ admits a generalized gradient at each point. The generalized
gradient coincides with the gradient at differentiable points and is a
convex set of points, called subgradients, at non-differentiable
points. The basic idea of subgradient methods is to generalize the 
methods for smooth problems by replacing the gradient by an arbitrary
subgradient. Algorithm 1 outlines the basic procedure of a generic
subgradient method. 

At differentiable points, direction finding generates a descent direction
$\vec{d}$ by exploiting the fact that the direction opposite to the gradient
of $f$ is locally the steepest descent direction. At non-differentiable
points, direction finding amounts in generating an arbitrary
subgradient. The problem is that a subgradient at a non-differentiable point 
is not necessarily a direction of descent. But according to Rademacher's
Theorem, the set of non-differentiable points is a
set of Lebesgue measure zero. Line search determines a step size $\eta_* >
0$ with which the current solution $\vec{x}^t$ is moved along direction
$\vec{d}$ in the updating step. Subgradient methods use predetermined step
sizes $\eta_{t,i}$, instead of some efficient univariate
smooth optimization method or polynomial interpolation as in gradient
descent methods. One reason for this is that a subgradient determined in the
direction finding step is not necessarily a direction of descent. Thus, the
viability of subgradient methods depend critically on the sequence of step
sizes. Updating moves the current solution $\vec{x}^t$ to the next
solution $\vec{x}^t + \eta_*\vec{d}$. Since the subgradient method is not a
descent method, it is common to keep track of the best point found so far,
i.e., the one with smallest function value. For more details on subgradient
methods and more advanced techniques to minimize locally Lipschitz
functions, we refer to \cite{Maekelae92}.

\begin{figure}[t]
\centering
\begin{algorithm}{Generic Subgradient Method}
\begin{small}
01 \atab{1} set $t := 0$ and choose starting point $\vec{x}^t \in \S{X}$\\
02 \atab{1} \textbf{repeat} \\
03 \atab{3} \textsc{Direction finding:}\\
04 \atab{5} determine $\vec{d} \in \S{X}$ and $\eta > 0$ such that
$f(\vec{x}^t + \eta \vec{d}) < f(f(\vec{x}^t)$\\[0.5ex] 
05 \atab{3} \textsc{Line search:}\\
06 \atab{5} find step size $\eta_* > 0$ such that $\eta_*\approx
\arg\min_{\eta > 0} f(\vec{x}^t + \eta \vec{d})$ \\[0.5ex] 
07 \atab{3} \textsc{Updating:}\\
08 \atab{5} set $\vec{x}^{t+1} := \vec{x}^t + \eta_* \vec{d}$\\[0.5ex] 
09 \atab{3} set $t := t+1$\\
10 \atab{1} \textbf{until} some termination criterion is satisfied
\end{small}
\end{algorithm}
\end{figure}

Several different subgradient methods for approximating a sample mean have been suggested \cite{Jain09b}. For extending k-means to the domain of graphs, we have chosen the incremental arithmetic mean (\verb|IAM|) method. In an empirical comparison of $8$ different subgradient methods \cite{Jain09b}, \verb|IAM| performed best with respect to computation time and was ranked third with respect to solution quality. In addition, \verb|IAM| best trades computation time and solution quality. For this reason, we consider \verb|IAM| as a good candidate for recomputing the centroids of the k-means clusters. 

\paragraph*{IAM -- Incremental Arithmetic Mean.}
The elementary incremental subgradient method randomly chooses a sample graph $X^t$ from $\S{S_T}$ at each iteration $t$ and updates the estimates $\vec{y}^{t} \in Y^{t}$ of the vector representations of a sample mean according to the formula
\[
\vec{y}^{t+1} =  \vec{y}^{t} - \eta^t\args{\vec{y}^{t} - \vec{x}^t}, 
\]
where $\eta^t$ is the step size and $(\vec{x}^t, \vec{y}^{t})$ is an optimal alignment. 

As a special case of the incremental subgradient algorithm, the incremental arithmetic mean method emulates the incremental calculation of the standard sample mean. First the order of the sample graphs from $\S{S_T}$ is randomly permuted. Then a sample mean is estimates according to the formula
\begin{align*}
\vec{y}^1 &= \vec{x}^1 \\
\vec{y}^i &= \frac{i-1}{i}\,\vec{y}^{i-1} + \frac{1}{i}\,\vec{x}^i \qquad
\mbox{for } 1 < i \leq N
\end{align*}
where $\args{\vec{x}i, \vec{y}^{i-1}}$ are optimal alignments for all $1 < i \leq N$. The graph $Y$ represented by the vector $\vec{y}^N$ is an approximation of a sample mean of $\S{S_T}$. In general, $Y$ is not an optimal solution of problem $(P)$. This procedure is inspired by Theorem 1.3 and requires only one iteration through the sample. The \verb|IAM| method requires $m-1$ distance calculations for approximating a sample mean. The solution quality depends on the order of selecting the sample graphs from $\S{S_T}$.

\subsection{The k-Means Algorithm for Graphs}

Having a graph metric and a concept of sample mean, we are now in the position, to extend the k-means algorithm to structure spaces $\S{X_T}$ over some Euclidean space $\S{X}$. We assume that $D$ is a distance metric induced by an Euclidean metric on $\S{X}$.  Now suppose that $\S{S_T} = \cbrace{X_{1}, \ldots, X_{N}}$ is a training sample of $N$ graphs drawn from $\S{X_T}$. We replace the standard cluster objective $J$ by 
\[
J_{\S{T}}\args{\vec{M},\S{Y_T} \,|\, \S{S_T}} = \sum_{j=1}^{k}\sum_{i=1}^{N} m_{ij} D\args{X_{i}, Y_{j}}{^2},
\]
where $\S{Y_T} = \cbrace{Y_1, \ldots, Y_k}$ is a set of $k$ centroids from $\S{X_T}$ and $\vec{M} = (m_{ij})$ is a membership matrix defining a partition of the set $\S{S_T}$.

Given a membership matrix $\vec{M}$, the cluster objective $J_{\S{T}}(.\,|\, \vec{M}, \vec{X})$ is no longer differentiable as a function of the centroids $\S{Y_T}$. But as shown in \cite{Jain08,Jain09b}, the objective  $J_{\S{T}}(.\,|\, \vec{M}, \vec{X})$ is locally Lipschitz and therefore differentiable as a function of $\S{Y_T}$ for almost all graphs. The $k$ centroids that minimize the cluster objective $J_{\S{T}}\args{.\,|\, \vec{M}, \vec{X_T}}$ are the structural versions of the sample mean
\[
Y_j = \arg\min_{Y \in \S{X_T}} F(Y) = \frac{1}{2} \sum_{i=1}^{N} m_{ij} D\argsS{X_i, Y}{^2}.
\]
Hence, we can easily extend Algorithm \ref{alg:k-means} to minimize the cluster objective $J_{\S{T}}$. Algorithm \ref{alg:k-means-s} describes the basic procedure of the k-means algorithm for structure spaces independent of the particular choice of method to minimize the objective $F$ of a sample mean. Similarly as for vectors, $\S{C}(Y)$ denotes the cluster associated with centroid $Y\in \S{Y_T}$. 

\begin{figure}[t]
\centering
\begin{algorithm}{K-Means Algorithm for Structure Spaces}\label{alg:k-means-s}
01 \atab{1} choose initial centroids $Y_{1}, \ldots, Y_{k}\in \S{X_T}$\\
02 \atab{1} \textbf{repeat} \\
03 \atab{3} assign each $X \in \S{S_T}$ to its closest centroid $Y_X = \arg\min_{Y\in\S{Y_T}} D\!\argsS{X, Y}{^2}$\\
04 \atab{3} recompute each $Y \in \S{Y_T}$ as a sample mean of all graphs from $\S{C}(Y)$\\
05 \atab{1} \textbf{until} some termination criterion is satisfied
\end{algorithm}
\end{figure}

In each iteration of the structural version of k-means requires $kN$ distance calculations to assign each pattern graph to a centroid and at least additional $O(N)$ distance calculations for recomputing the centroids using the incremental arithmetic mean subgradient method.  This gives a total of at least $O(kN + N)$ distance calculations in each iteration of Algorithm \ref{alg:k-means-s}.

\section{Elkan's k-Means for Graphs}

In this section we extend Elkan's k-means \cite{Elkan03} from vectors to graphs. 

\introskip

\begin{figure}[t]
\begin{center}
\begin{algorithm}{Elkan's k-Means Algorithm for Structure Spaces}\label{alg:elkan-k-means-s}
01 \atab{1} choose set $\S{Y_T} = \cbrace{Y_{1}, \ldots, Y_{k}}$ of initial centroids \\
02 \atab{1} set $l(X, Y) = 0$ for all $X \in \S{S_T}$ and for all $Y\in\S{Y_T}$\\ 
03 \atab{1} set $u(X) = \infty$ for all $X\in \S{S_T}$\\
04 \atab{1} randomly assign each $X\in \S{S_T}$ to a centroid $Y_X \in \S{Y_T}$\\
05 \atab{1} \textbf{repeat} \\
06 \atab{3} compute $D\args{Y, Y'}$ for all centroids $Y, Y' \in \S{Y_T}$\\
07 \atab{3} \textbf{for} each $X \in \S{S_T}$ and $Y \in \S{Y_T}$ \textbf{do}\\
08 \atab{5} \textbf{if} $Y$ is a candidate centroid for $X$\\
09 \atab{7} \textbf{if} $u(X)$ is out-of-date \\
10 \atab{9} update $u(X) = D\args{X, Y_X}$\\
11 \atab{9} update $l\args{X, Y_X} = l(X)$\\
12 \atab{7} \textbf{if} $Y$ is a candidate centroid for $X$\\
13 \atab{9} update $l\args{X,Y} = D\args{X, Y}$\\
14 \atab{9} \textbf{if} $l\args{X, Y} \leq u\args{X}$\\
15 \atab{11} update $u(X) = l\args{X, Y}$\\
16 \atab{11} replace $Y_X = Y$\\
17 \atab{3} recompute mean $\check{Y}$ of cluster $\S{C}(Y)$ for all $Y\in \S{Y_T}$\\
18 \atab{3} compute $\delta(Y) = D\args{Y, \check{Y}}$  for all $Y\in \S{Y_T}$\\
19 \atab{3} set $u(X) = u(X) + \delta\args{Y_X}$ for all $X\in \S{X_T}$\\
20 \atab{3} set $l(X,Y) = \max\cbrace{l(X,Y) - \delta\args{Y}, 0}$ for all $X\in \S{X_T}$ and for all $Y\in \S{Y_T}$\\
21 \atab{3} replace $Y$ by $\check{Y}$ for all $Y \in \S{Y_T}$\\
22 \atab{1} \textbf{until} some termination criterion is satisfied.
\end{algorithm}
\end{center} 
\vspace{-5ex}
\noindent
\begin{footnotesize}
\noindent
\emph{Remark}: 
\vspace{-0.2cm}
\begin{enumerate}
\item
Setting the value a variable such as $l_*(\check{Y}, \check{Y}')$ and $u(X)$ implicitly declares the value of that variable as out-of-date. Updating those variables declares the value as up-to-date. 
\item
The condition in line \texttt{08} and \texttt{12} is only redundant if the upper bound $u(X)$ is up-to-date.
\end{enumerate}
\end{footnotesize}  
\hrule
\end{figure}

Frequent evaluation of NP-hard graph distances dominates the computational cost of k-means for graphs. Accelerating k-means therefore aims at reducing the number of graph distance calculations. In \cite{Elkan03}, Elkan suggested an accelerated formulation of the standard k-means algorithm for vectors exploiting the triangle inequality of the underlying distance metric. Since the distance function $D$ on $\S{X_T}$ induced by an Euclidean metric is also a metric \cite{Jain09c}, we can transfer Elkan's k-Means acceleration from Euclidean spaces to $\S{T}$-spaces.

To extend Elkan's k-Means acceleration to $\S{T}$-spaces, we assume that $X \in \S{S_T}$ is a pattern graph and $Y, Y' \in \S{Y_T}$ are centroids. As before, by $Y_X$ we denote the centroid the pattern graph $X$ is assigned to. Elkan's acceleration is based on two observations:
\begin{enumerate}
\item From the triangle inequality of a metric follows
\begin{align}
\label{eq:triangle01}
u(X) \leq \frac{1}{2}D\args{Y_X, Y} \;\Rightarrow\; D\args{X, Y_X} \leq D\args{X, Y},
\end{align}
where $u(X) \geq D\args{X, Y_X}$ denotes an upper bound of the distance $D\args{X, Y_X}$. 
\item We have
\begin{align}
\label{eq:triangle02}
u(X) \leq l(X, Y) \;\Rightarrow\; D\args{X, Y_X} \leq D\args{X, Y},
\end{align}
where $l(X, Y) \leq D\args{X, Y}$ denotes a lower bound of the distance $D\args{X, Y}$.
\end{enumerate}
As an immediate consequence, we safely can avoid to calculate a distance $D\args{X, Y}$ between a pattern graph $X$ and an arbitrary centroid $Y$ if at least one of the following conditions is satisfied
\begin{description}
\item[$(C_1)$] $Y = Y_X$
\item[$(C_2)$] $u(X) \leq \frac{1}{2}D\args{Y_X, Y}$
\item[$(C_3)$] $u(X) \leq l(X, Y)$
\end{description}
We say, $Y$ is a \emph{candidate centroid} for $X$ if all conditions $(C_1)$-$(C_3)$ are violated. Conversely, if $Y$ is not a candidate centroid for $X$, then either condition $(C_2)$ or condition $(C_1)$ is satisfied. From the inequalities (\ref{eq:triangle01}) and (\ref{eq:triangle01}) follows that $Y$ can not be a centroid closest to $X$. Therefore, it is not necessary to calculate the distance $D(X, Y)$. In the case that $Y_X$ is the onliest candidate centroid for $X$ all distance calculations $D(X, Y)$ with $Y\in \S{Y_T}$ can be skipped and $X$ must remain assigned to $Y_X$.

Now suppose that $Y \neq Y_X$ is a candidate centroid for $X$. Then we apply the technique of "delayed (distance) evaluation". We first test whether the upper bound $u(X)$ is out-of-date, i.e.\ if $u(X) \gneqq D\args{X, Y_X}$. If $u(X)$ is out-of-date we improve the upper bound by setting $u(X) = D\args{X, Y_X}$. Since improving $u(X)$ might eliminate $Y$ as being a candidate centroid for $X$, we again check conditions $(C_2)$ and $(C_3)$. If both conditions are still violated despite the updated upper bound $u(X)$, we have the following situation
\begin{align*}
u(X) = D\args{X, Y_X} &> \frac{1}{2}D\args{Y_X, Y}
u(X) = D\args{X, Y_X} &> l(X, Y).
\end{align*}
Since the distances on the left and right hand side of the inequality of condition $(C_2)$ are known, we may conclude that the situation for condition $(C_2)$ can not be altered. Therefore, we re-examine condition $(C_3)$ by calculating the distance $D(X, Y)$ and updating the lower bound $l(X) = D(X, Y)$. If condition $(C_3)$ is still violated, we have
\[
u(X) = D\args{X, Y_X} > D(X, Y) = l(X, y). 
\]
This implies that $X$ is closer to centroid $Y$ than to $Y_X$ and therefore has to be assigned to centroid $Y$. 

Crucial for avoiding distance calculations are good estimates of the lower and upper bounds $l(X, Y)$ and $u(X)$ in each iteration. For this, we compute the change $\delta(Y)$ of each centroid $Y$ by the distance 
\[
\delta(Y) = D(Y, \check{Y}), 
\]
where $\check{Y}$ is the recomputed centroid of cluster $\S{C}(Y)$. Based on the triangle inequality, we set the bounds according to the following rules
\begin{align}
\label{eq:rule-for-lowerbound}
l(X, Y) &= \max\cbrace{l(X,Y) - \delta(Y), 0}\\
\label{eq:rule-for-upperbound}
u(X)    &= u(X) + \delta\args{Y_X}.
\end{align}
In addition, $u(X)$ is then declared as out-of-date.\footnote{In the original formulation of Elkan's algorithm for feature vectors, the upper bounds $u(X)$ are declared as out-of-date regardless of the value $\delta(Y)$.}
Both rules guarantee that $l(X, Y)$ is always a lower bound of $D\args{X, Y}$ and $u(X)$ is always an upper bound of $D\args{X, Y_X}$. 

Algorithm \ref{alg:elkan-k-means-s} presents a detailed description of Elkan's k-means algorithm for graphs. During each iteration, $k(k-1)/2$ pairwise distances between all centers must be recomputed (Algorithm \ref{alg:elkan-k-means-s}, line \verb|07|). Recomputing the centroids using incremental arithmetic mean (see Section 3.2) requires additional $O(N)$ distance calculations (Algorithm \ref{alg:elkan-k-means-s}, line \verb|19|-\verb|20|). To update the lower and upper bounds, $k$ distances between the current and the new centroids must be calculated (Algorithm \ref{alg:elkan-k-means-s}, line \verb|21|-\verb|25|). This gives a minimum of $O\args{N + k^2}$ distance calculations at each iteration ignoring the delayed distance evaluations in line \verb|09|-\verb|18| of Algorithm \ref{alg:elkan-k-means-s}. As the centroids converge, one would expect that the partition of the training sample becomes more and more stable, which results in a decreasing number of delayed distance evaluations. 

\section{Experiments}
\begin{table}[t]
\begin{tabular}{lcccccc}
\hline
\hline
data set & \#(graphs)& \#(classes) & avg(nodes) & max(nodes) & avg(edges) & max(edges) \\
\hline
letter & 750 & 15 & 4.7 & 8 & 3.1 & 6\\
grec & 528 & 22 & 11.5 & 24 & 11.9 & 29\\
fingerprint & 900 & 3 & 8.3 & 26 & 14.1 & 48\\
molecules & 100 & 2 & 24.6 & 40 & 25.2 & 44\\
\hline
\hline
\end{tabular}
\vspace{1ex}
\caption{Summary of main characteristics of the data sets.}
\label{tab:characteristics}
\end{table}

This section reports the results of running k-means and Elkan's k-means on four graph data sets. 

\subsection{Data.} 
We selected four data sets described in \cite{Riesen08}. The data sets are publicly available at \cite{IAMGDB}. Each data set is divided into a training, validation, and a test set. In all four cases, we considered data from the test set only. The description of the data sets are mainly excerpts from \cite{Riesen08}. Table \ref{tab:characteristics} provides a summary of the main characteristics of the data sets.

\begin{figure}[hp]
\centering
\includegraphics[width=0.4\textwidth]{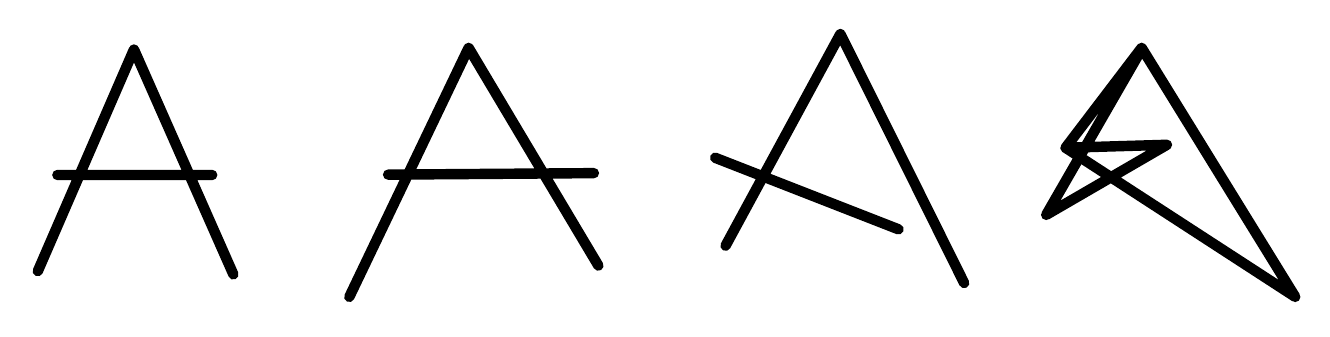}
\caption{Example of letter drawings: Prototype of letter A and distorted copies generated by imposing low, medium, and high distortion (from left to right) on prototype A.}
\label{fig:letters}
\end{figure}

\paragraph*{Letter Graphs.}
We consider all $750$ graphs from the test data set representing distorted letter drawings from the Roman alphabet that consist of straight lines only (A, E, F, H, I, K, L, M, N, T, V, W, X, Y, Z). The graphs are uniformly distributed over the $15$ classes (letters). The letter drawings are obtained by distorting prototype letters at low distortion level. Lines of a letter are represented by edges and ending points of lines by vertices. Each vertex is labeled with a two-dimensional vector giving the position of its end point relative to a reference coordinate system. Edges are labeled with weight $1$. Figure \ref{fig:letters} shows a prototype letter and distorted version at various distortion levels. 

\begin{figure}[hp]
\centering
\includegraphics[width=0.5\textwidth]{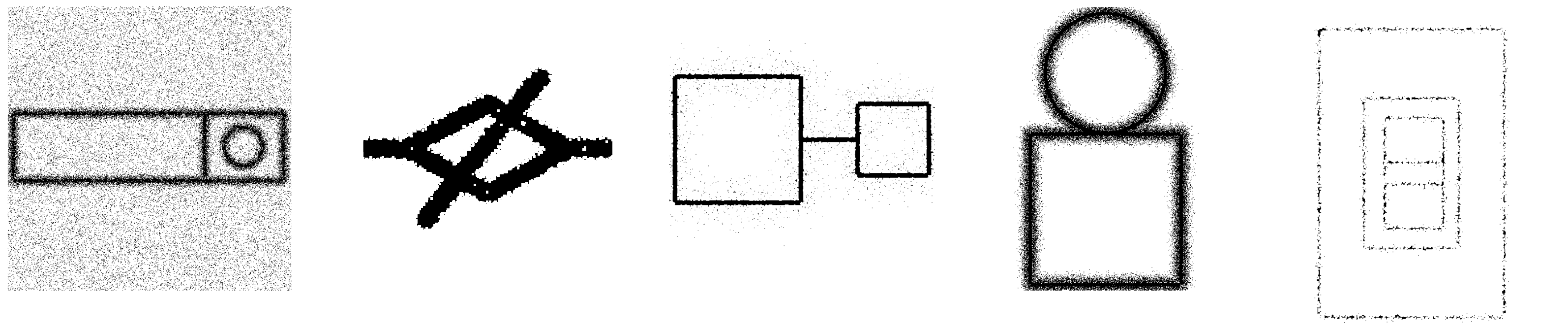}
\caption{GREC symbols: A sample image of each distortion level}
\label{fig:grec}
\end{figure}

\paragraph*{GREC Graphs.}
The GREC data set \cite{Dosch06} consists of graphs representing symbols from architectural and electronic drawings. We use all $528$ graphs from the test data set uniformly distributed over $22$ classes. The images occur at five different distortion levels. In Figure \ref{fig:grec} for each distortion level one example of a drawing is given. Depending on the distortion level, either erosion, dilation, or other morphological operations are applied. The result is thinned to obtain lines of one pixel width. Finally, graphs are extracted from the resulting denoised images by tracing the lines from end to end and detecting intersections as well as corners. Ending points, corners, intersections and circles are represented by vertices and labeled with a two-dimensional attribute giving their position. The vertices are connected by undirected edges which are labeled as line or arc. An additional attribute specifies the angle with respect to the horizontal direction or the diameter in case of arcs. 

\begin{figure}[hp]
\centering
\includegraphics[width=0.5\textwidth]{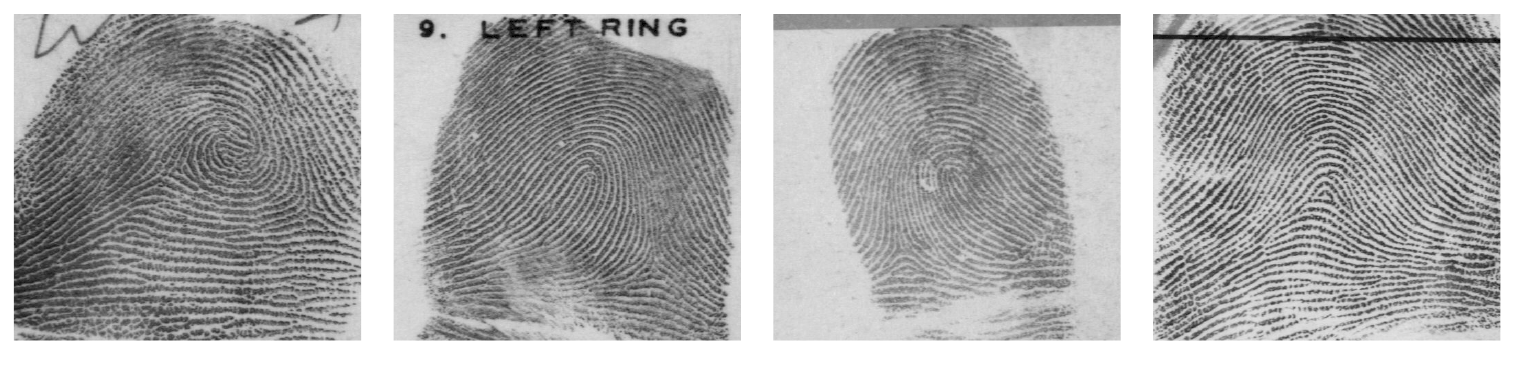}
\caption{Fingerprints: (a) Left (b) Right (c) Arch (d) Whorl. Fingerprints of class arch are not considered.}
\label{fig:fingerprints}
\end{figure}

\paragraph*{Fingerprint Graphs.}
We consider a subset of $900$ graphs from the test data set representing fingerprint images of the NIST-4 database \cite{Watson92}. The graphs are uniformly distributed over three classes \emph{left}, \emph{right}, and \emph{whorl}. A fourth class (\emph{arch}) is excluded in order to keep the data set balanced. Fingerprint images are converted into graphs by filtering the images and extracting regions that are relevant \cite{Neuhaus05}. Relevant regions are binarized and a noise removal and thinning procedure is applied. This results in a skeletonized representation of the extracted regions. Ending points and bifurcation points of the skeletonized regions are represented by vertices. Additional vertices are inserted in regular intervals between ending points and bifurcation points. Finally, undirected edges are inserted to link vertices that are directly connected through a ridge in the skeleton. Each vertex is labeled with a two-dimensional attribute giving its position. Edges are attributed with an angle denoting the orientation of the edge with respect to the horizontal direction. Figure \ref{fig:fingerprints} shows fingerprints of each class. 

\paragraph*{Molecules.}
The mutagenicity data set consists of chemical molecules from two classes (mutagen, non-mutagen). The data set was originally compiled by \cite{Kazius05} and reprocessed by \cite{Riesen08}. We consider a subset of $100$ molecules from the test data set uniformly distributed over both classes.
We describe molecules by graphs in the usual way: atoms are represented by vertices labeled with the atom type of the corresponding atom and bonds between atoms are represented by edges labeled with the valence of the corresponding bonds. We used a $1$-to-$k$ binary encoding for representing atom types and valence of bonds, respectively.

\subsection{General Experimental Setup}

In all experiments, we applied standard k-means for graphs (std) and Elkan's k-means for graphs (elk) to the aforementioned data sets using the following experimental setup:

\paragraph*{Setting of k-means algorithms.}
To initialize the k-means algorithms, we used a modified version of the "furthest first" heuristic \cite{Hochbaum85}. For each data set $\S{S}$, the first centroid $Y_1$ is initialized to be a graph closest to the sample mean of $\S{S}$. Subsequent centroids are initialized according to 
\[
Y_{i+1} = \arg \max_{X \in \S{S}} \min_{Y \in \S{Y}_i} D(X, Y),
\]
where $\S{Y}_i$ is the set of the first $i$ centroids chosen so far. We terminated each k-means algorithm after $3$ iterations without improvement of the cluster objective $J_{\S{T}}$.

\paragraph*{Graph distance calculations and optimal alignment.}
For graph distance calculations and finding optimal alignments, we applied 
a depth first search algorithm on the letter data set and the graduated assignment \cite{Gold96} on the grec, fingerprint, and molecule data set. 
The depth first search method guarantees to return optimal solutions and therefore can be applied to small graphs only. Graduated assignment returns approximate solutions. 

\paragraph*{Performance measures.}
We used the following measures to assess the performance of an algorithm on a dataset: (1) error (value of the cluster objective $J_{\S{T}}$), (2) classification accuracy, (3) silhouette index, and (4) number of graph distance calculations.  

The silhouette index is a cluster validation index taking values from $[-1, 1]$. Higher values indicate a more compact and well separated cluster structure. For more details we refer to Appendix \ref{app:silhouette} and \cite{Theodoridis09}. Elkan's k-means and graph-vector reduction k-means incur computational overhead to create and update auxiliary data structures and to compute Euclidean distances. This overhead is negligible compared to the time spent on graph distance calculations. Therefore, we report number of graph distance calculations rather than clock times as a performance measure for speed.

\subsection{Performance Comparison}

\begin{table}[tp]
\centering
\begin{footnotesize}
\begin{tabular}{l@{\qquad}l@{\qquad}l@{\qquad}c@{\quad}c}
\hline
\hline
data set    & k & measure & std & elk\\
\hline
\hline
letter      & 30 &                &          &     \\
            &    & error          & 11.6     & 11.5 \\
		        &    & accuracy       & 0.86     & 0.86 \\
            &    & silhouette     & 0.38     & 0.39 \\
            &    & iterations     & 21       & 13  \\
						&    & matchings  $\args{\times 10^3}$\\
						&    & \quad per iteration &  23.2 &   3.3 \\
						&    & \quad total         & 488.4  & 42.5 \\
						&    & speedup \\
						&    & \quad per iteration & 1.0 &  7.1  \\
						&    & \quad total         & 1.0 & 11.5 \\
\hline
grec			  & 33 &                &          &          \\
    			  &    & error          & 32.7     & 32.2    \\
            &    & accuracy       & 0.84     & 0.83    \\
            &    & silhouette     & 0.40     & 0.44     \\
            &    & iterations     & 11       & 11      \\
						&    & matchings  $\args{\times 10^3}$\\
						&    & \quad per iteration & 18.0  &  5.7  \\
						&    & \quad total         & 197.5 & 63.1  \\
						&    & speedup \\
						&    & \quad per iteration & 1.0 & 3.1  \\
						&    & \quad total         & 1.0 & 3.1  \\
\hline
fingerprint & 60 &                &          &         \\
            &    & error          & 1.88     & 1.70    \\
            &    & accuracy       & 0.81     & 0.82 \\
            &    & silhouette     & 0.32     & 0.31  \\
            &    & iterations     & 10        & 11      \\
						&    & matchings  $\args{\times 10^3}$\\
						&    & \quad per iteration & 54.9 & 4.8 \\
						&    & \quad total         & 549  & 52.4 \\
						&    & speedup \\
						&    & \quad per iteration & 1.0 & 11.5 \\
						&    & \quad total         & 1.0 & 10.5 \\
\hline
molecules   & 10 &                &          &      \\
            &    & error          & 27.6     & 27.2   \\
		        &    & accuracy       & 0.69     & 0.70    \\
            &    & silhouette     & 0.03     & 0.04    \\
            &    & iterations     & 13       & 13      \\
						&    & matchings $\args{\times 10^3}$\\
						&    & \quad per iteration &  1.1&  1.1\\
						&    & \quad total         & 14.3& 14.5\\
						&    & speedup \\
						&    & \quad per iteration & 1.0 &  0.94 \\
						&    & \quad total         & 1.0 &  0.94 \\
\hline
\hline
\end{tabular}
\end{footnotesize}
\vspace{2ex}
\caption{Results of different k-means clusterings on four data sets. Columns labeled with \emph{std}, \emph{elk}, and \emph{gvr} give the performance of standard k-means for graphs, Elkan's k-means for graphs, and graph-vector reduction k-means, respectively. Rows labeled \emph{matchings} give the number of distance calculations $\args{\times 10^3}$, and rows labeled \emph{speedup} show how many times an algorithm is faster than standard k-means for graphs.}
\label{tab:results}
\end{table}

We applied standard k-means (std) and Elkan's k-means (elk) to all four data sets in order to assess and compare their performance. The number $k$ of centroids as shown in Table \ref{tab:results} was chosen by compromising a satisfactory classification accuracy against the silhouette index. For each data set $5$ runs of each algorithm were performed and the best cluster result selected. 

Table \ref{tab:results} summarizes the results. The first observation to be made is that the solution quality of std and elk is comparable with respect to error, classification accuracy, and silhouette index. Deviations are due to the non-uniqueness of the sample mean and the approximation errors of the graduated assignment algorithm. The second observation to be made from Table \ref{tab:results} is that elk outperforms std with respect to computation time on the letter, grec, and fingerprint data set. On the molecule data set, std and elk have comparable speed performance. Remarkably, elk requires slightly more distance calculations than std. 

Contrasting the silhouette index and the dimensionality of the data to the speedup factor gained by elk, we make the following observation: First, the silhouette index for the letter, grec, and fingerprint data set are roughly comparable and indicate a cluster structure in the data, whereas the silhouette index for the molecule data set indicates almost no compact and homogeneous cluster structure. Second, the dimensionality of the vector representations is largest for molecule graphs, moderate for grec graphs, and relatively low for letter and fingerprint graphs. Thus, the speedup factor of elk and gvr apparently decreases with increasing dimensionality and decreasing cluster structure. This behavior is in line with findings in high-dimensional vector spaces \cite{Elkan03}. According to \cite{Moore00}, there will be little or no acceleration in high dimensions if there is no underlying structure in the data. This view is also supported by theoretical results from computational geometry \cite{Indyk99}.

\subsection{Speedup vs. Number $k$ of Centroids}

\begin{table}[hp]
\centering
\begin{footnotesize}
\begin{tabular}{l@{\qquad}l@{\qquad}l@{\qquad}c@{\quad}c}
\hline
\hline
data set    & k & measure & std & elk \\
\hline
\hline
letter      & 4 &                &          &          \\
(A, E, F, H)&    & error         & 6.9      & 7.0     \\
		        &    & accuracy      & 0.61     & 0.60  \\
            &    & silhouette    & 0.26     & 0.25    \\
            &    & iterations    & 14       & 15  \\
						&    & matchings  $\args{\times 10^2}$\\
						&    & \quad per iteration &  10.0&  4.4 \\
						&    & \quad total         & 140.0& 65.5  \\
						&    & speedup \\
						&    & \quad per iteration & 1.0 &  2.3  \\
						&    & \quad total         & 1.0 &  2.1 \\
\hline
letter			& 8 &                &         &          \\
(A, E, F, H)&    & error         & 4.2     & 4.2     \\
            &    & accuracy      & 0.82    & 0.82    \\
            &    & silhouette    & 0.30    & 0.30   \\
            &    & iterations    & 13      & 14  \\
						&    & matchings  $\args{\times 10^2}$\\
						&    & \quad per iteration &  18.0 &   5.1 \\
						&    & \quad total         & 234.0 &  71.3\\
						&    & speedup \\
						&    & \quad per iteration & 1.0 & 3.5 \\
						&    & \quad total         & 1.0 & 3.3 \\
\hline
letter      & 12 &                &          &         \\
(A, E, F, H)&    & error          & 2.7     & 2.7   \\
            &    & accuracy       & 0.94     & 0.94 \\
            &    & silhouette     & 0.31     & 0.30 \\
            &    & iterations     & 16        & 18   \\
						&    & matchings  $\args{\times 10^2}$\\
						&    & \quad per iteration & 26.0 &  5.5  \\
						&    & \quad total         & 416  & 98.7 \\
						&    & speedup \\
						&    & \quad per iteration & 1.0 & 4.2 \\
						&    & \quad total         & 1.0 & 4.7 \\
\hline
letter      & 16 &                &          &           \\
(A, E, F, H)&    & error          & 2.4     & 2.4   \\
            &    & accuracy       & 0.94     & 0.94 \\
            &    & silhouette     & 0.22     & 0.23   \\
            &    & iterations     & 16        & 16     \\
						&    & matchings  $\args{\times 10^2}$\\
						&    & \quad per iteration & 34.0 &  6.7  \\
						&    & \quad total         & 544.0  & 107.8\\
						&    & speedup \\
						&    & \quad per iteration & 1.0 & 5.0 \\
						&    & \quad total         & 1.0 & 6.1 \\
\hline
\hline
\end{tabular}
\end{footnotesize}
\vspace{2ex}
\caption{Results of k-means clusterings on a subset of the letter graphs (A, E, F, H) for four different values of $k = 4, 8, 12, 16$. Shown are the average values of the performance measures averaged over $10$ runs.}
\label{tab:results2}
\end{table}

 \begin{table}[hp]
\centering
\begin{footnotesize}
\begin{tabular}{l@{\qquad}l@{\qquad}l@{\qquad}c@{\quad}c}
\hline
\hline
data set    & k & measure & std & elk \\
\hline
\hline
fingerprints& 3  \\
            &    & error         & 41.3     & 41.5     \\
		        &    & accuracy      & 0.59     & 0.59   \\
            &    & silhouette    & 0.20     & 0.21    \\
            &    & iterations    & 7        & 7  \\
						&    & matchings  $\args{\times 10^2}$\\
						&    & \quad per iteration & 12.0&  9.5 \\
						&    & \quad total         & 84.0& 66.8 \\
						&    & speedup \\
						&    & \quad per iteration & 1.0 &  1.3 \\
						&    & \quad total         & 1.0 &  1.3\\
\hline
fingerprints& 15 \\
            &    & error         & 8.1     & 6.8    \\
            &    & accuracy      & 0.64    & 0.64   \\
            &    & silhouette    & 0.32    & 0.33   \\
            &    & iterations    & 10      & 9       \\
						&    & matchings  $\args{\times 10^2}$\\
						&    & \quad per iteration &  48.0 &   15.4\\
						&    & \quad total         & 480.0 &  138.4  \\
						&    & speedup \\
						&    & \quad per iteration & 1.0 & 3.1  \\
						&    & \quad total         & 1.0 & 3.5  \\
\hline
\hline
\end{tabular}
\end{footnotesize}
\vspace{2ex}
\caption{Results of k-means clusterings on a subset of the fingerprints graphs for two different values of $k = 3$ and $k = 15$. Shown are the average values of the performance measures averaged over $10$ runs.}
\label{tab:results3}
\end{table}

In this experiment we investigate how the speedup factor of elk depends on the number $k$ of centroids. For this, we restricted to subsets of the letter and fingerprint data sets. We selected $200$ graphs uniformly distributed over the four classes A, E, F, and H. From the fingerprint data set we compiled a subset of $300$ graphs uniformly distributed over all three classes. For each chosen number $k$ of centroids $10$ runs of each algorithm were conducted and the average of all performance measures was taken.  The number $k$ is shown in Table \ref{tab:results2} for letter graphs and Table \ref{tab:results3} for fingerprints graphs. 

From the results shown in Table \ref{tab:results2} and \ref{tab:results3}, we see that the speedup factor slowly increases with increasing number $k$ of centroids. The results confirm that std and elk perform comparable with respect to solution quality for varying $k$. As an aside, all k-means algorithms for graphs exhibit a well-behaved performance in the sense that subgradient methods applied to the nonsmooth cluster objective $J_{\S{T}}$ indeed minimize $J_{\S{T}}$ in a reasonable way as shown by the decreasing error for increasing $k$.

\section{Conclusion}

We extended Elkan's k-means from vectors to graphs. Elkan's k-means exploits the triangle inequality to avoid graph distance calculations. Experimental results show that standard and Elkan's k-means for graphs perform equally with respect to solution quality, but Elkan's k-means outperforms standard k-means with respect to speed if there is a cluster structure in the data. The speedup factor of both accelerations increases slightly with the number $k$ of centroids. This contribution is a first step in accelerating clustering algorithms that directly operate in the domain of graphs. Future work aims at accelerating incremental clustering methods.

\begin{appendix}
\section{The Silhouette Index}\label{app:silhouette}
Suppose that $\S{S} = \cbrace{X_1, \ldots, X_m}$ is a sample of $m$ patterns. 
Let $\S{C} = \cbrace{\S{C}_1, \ldots, \S{C}_k}$ be a partition of $\S{S}$ consisting of $k$ disjoint clusters with 
\[
\S{S} = \bigcup_{i=1}^{k}\S{C}_i.
\] 
We assume that $D$ is the underlying distance function defined on $\S{S}$. The distance between two subsets $\S{U}, \S{U}' \subseteq \S{S}$ is defined by 
\[
D\args{\S{U}, \S{U}'} = \min\cbrace{D\args{X, X'} \,:\, X \in \S{U}, X'\in \S{U}'}.
\]
If $\S{U} = \cbrace{X}$ consists of a singleton, we simply write $D\args{X, \S{U}'}$ instead of $D\args{\cbrace{X}, \S{U}'}$. 

Let
\[
D_{\text{avg}}\args{X, \S{U}} 
\]
denote the average distance between pattern $X \in \S{S}$ and subset $\S{U} \subseteq \S{S}$. Suppose that pattern $X_i \in \S{S}$ is a member of cluster $\S{C}_{m(i)} \in \S{C}$. By $\S{C}'_{m(i)}$ we denote the set $\S{C}_{m(i)} \setminus\cbrace{X_i}$. For each pattern $X_i \in \S{S}$ let
\[
a_i = D_{\text{avg}}\args{X, \S{C}'_{m(i)}}
\] 
be the average distance between pattern $X_i$ and subset $\S{C}'_{m(i)}$. By 
\[
b_i = \min_{j \neq m(i)} D_{\text{avg}}\args{X_i, \S{C}_j}
\]
we denote the minimum average distance between pattern $X_i$ and all clusters from $\S{C}$ not containing $X_i$. The \emph{silhouette width} of $X_i$ is defined as
\[
s_i = \frac{b_i - a_i}{\max\args{b_i, a_i}}. 
\] 
The \emph{silhouette of cluster} $\S{C}_j \in \S{C}$ is given by
\[
S_j = \frac{1}{\abs{\S{C}_j}} \sum_{i:X_i\in\S{C}_j} s_i.
\]
The \emph{silhouette index} is then defined as the average of all cluster  silhouettes 
\[
\mathfrak{S} = \frac{1}{k} \sum_{j=1}^{k} S_j.
\]
\end{appendix}

\bibliographystyle{splncs}

\end{document}